\documentclass[letters]{fcs}
\usepackage{layout}
\usepackage{subfig}

\usepackage{amsmath,amsfonts,bm}

\def\eqref#1{equation~\ref{#1}}

\def\1{\bm{1}}

\DeclareMathAlphabet{\mathsfit}{\encodingdefault}{\sfdefault}{m}{sl}
\SetMathAlphabet{\mathsfit}{bold}{\encodingdefault}{\sfdefault}{bx}{n}

\usepackage{hyperref}
\usepackage{url}
\usepackage{algorithm}
\usepackage{algorithmic}
\usepackage{graphicx}
\usepackage{amsmath}
\usepackage{amsthm}
\usepackage{amssymb}

\title{Policy Improvement with Style-Specific Demonstrations}

\author[1]{Lingfeng Li}
\author[1]{Yunlong Lu}
\author[1]{Yongyi Wang}
\author[1,+]{Wenxin Li}
\address[1]{School of Computer Science, Peking University, Beijing 100871, China}

\corremail{lwx@pku.edu.cn}
\fcssetup{
  received = {month dd, yyyy},
  accepted = {month dd, yyyy},
}

\begin{document}

\section{Introduction}

Games benefit from AI agents with diverse play styles. In MOBAs and hero shooters, where characters possess fundamentally different mechanics---from aggressive assassins to defensive supports---authentic stylistic variety is essential for immersive player experiences. Reinforcement learning (RL) has driven game AI to superhuman performance across board games, computer games, and esports~\cite{silver2018general}, yet these methods optimize exclusively for reward maximization and disregard behavioral style. Quality-Diversity optimization~\cite{10.1145/2001576.2001606} and population-based RL~\cite{parkerholder2020effectivediversitypopulationbased} can generate varied policies, but at huge computational cost and with limited control over individual agent styles.

Learning from Demonstration (LfD) offers an alternative paradigm. Pure imitation learning methods such as GAIL~\cite{DBLP:journals/corr/HoE16} learn from expert demonstrations but rarely surpass the demonstrator and suffer from adversarial training instability. In contrast, RL-with-demonstrations approaches integrate offline data into online RL; among these, DQfD~\cite{DBLP:journals/corr/HesterVPLSPSDOA17} is most relevant to discrete-action game settings, using explicit supervised losses to guide policy learning. MPPO differs by replacing explicit imitation objectives with an implicit behavioral constraint---demonstration data is mixed directly into the PPO surrogate objective under a single loss, naturally handling suboptimal demonstrations while enabling full-episode GAE via compact seed-based trajectories.

We address the practical problem of improving the proficiency of suboptimal, stylized agents while preserving their play styles. MPPO unifies online interaction data and offline demonstration samples within a PPO-style clipped surrogate objective, where a mixing coefficient $\beta$ governs the trade-off between performance improvement and style preservation. Our seed-based demonstration storage mechanism reduces storage by over 98\% in our test scenarios.

We evaluate MPPO across three environments of increasing complexity: Blackjack, Maze Navigation, and MCR Mahjong. Across all environments, MPPO agents consistently surpass their suboptimal demonstrators in proficiency while maintaining substantially smaller policy distances, measured by $D_{policy}$, a total-variation-based metric we introduce, than pure online PPO. In Mahjong, one MPPO agent surpasses the top-ranked bot on Botzone's Elo ranking list~\cite{10.1145/3197091.3197099}. Our key contributions are:
\begin{itemize}[leftmargin=*,noitemsep,nolistsep]
\item MPPO, which leverages suboptimal demonstrator data to improve policy proficiency while constraining deviation from the demonstrator's behavior through implicit, data-driven constraints;
\item A unified PPO-style loss formulation connecting online and offline data under mixed state distributions;
\item A seed-based demonstration replay mechanism enabling full-episode advantage estimation with 98\%+ storage reduction;
\item A multi-environment benchmark with stylized bots and the $D_{policy}$ metric for quantifying play-style similarity.
\end{itemize}

\section{Proposed Algorithm}
\label{gen_inst}

We formulate game environments as standard Markov Decision Processes (MDP) $\mathcal{M} = \langle \mathcal{S}, \mathcal{A}, R, \mathcal{P}, \gamma, \rho_0\rangle$, where $\mathcal{S}$ and $\mathcal{A}$ are the observable state and action spaces, $R(s,a)$ is the reward function, $\gamma \in (0,1)$ is the discount factor, and $\rho_0$ is the initial state distribution. Policy $\pi(a_t|s_t)$ is the distribution of actions conditioned on states, and $\mathcal{P}(s'|s,a)$ is the transition distribution capturing both inherent environment stochasticity and the effects of other agents' hidden decisions. The horizon $T$ denotes the maximum episode length. The performance of $\pi$ is $J(\pi) = \mathbb{E}_\pi \sum_{t=0}^{T}\gamma^t R(s_t,a_t)$, and the advantage function is $A_{\pi}(s,a) = Q_{\pi}(s,a) - V_{\pi}(s)$.

We introduce \textbf{M}ixed \textbf{P}roximal \textbf{P}olicy \textbf{O}ptimization (MPPO), which integrates data from both online interactions and offline demonstration datasets. Standard PPO~\cite{DBLP:journals/corr/SchulmanWDRK17} optimizes the clipped surrogate objective using data collected exclusively by the current policy $\pi_k$, driving the agent toward reward-maximizing behavior. When initialized from a stylized policy, this can rapidly erase the demonstrator's stylistic characteristics. To guide the policy toward higher reward while preventing excessive stylistic drift, we mix demonstration data directly into the PPO objective: a fraction $\beta$ of state-action tuples are sampled from a pre-collected demonstration dataset $D$, and the remainder $1-\beta$ from online interactions. The resulting MPPO loss is:
$$
 L^{\textit{MPPO}} = \beta \mathbb{E}_{s\sim\rho_D(\cdot)}[\text{min}(rA_{\pi_k}(s,a),\text{clip}(r, 1\!-\epsilon,1\!+\epsilon)A_{\pi_k}(s,a))]
$$
\begin{equation}
 + (1\!-\!\beta) \mathbb{E}_{s\sim\rho_{\pi_k}(\cdot)}[\text{min}(rA_{\pi_k}(s,a),\text{clip}(r, 1\!-\epsilon,1\!+\epsilon)A_{\pi_k}(s,a))] \label{eq:4}
\end{equation}
where $r = \frac{\pi_k'(a|s)}{\pi_k(a|s)}$ is the probability ratio and $\epsilon$ controls the per-update step size. For on-policy samples, $r$ forms the standard PPO clipping mechanism; for off-policy data, $r$ additionally serves as an importance sampling ratio correcting for the distribution shift between demonstrator data and the current policy.

The key design choice is that style preservation emerges from the training data distribution itself, rather than from an auxiliary imitation loss. By mixing demonstration trajectories directly into the PPO objective, the policy is trained on a blend of the current policy's exploratory behavior and the demonstrator's characteristic actions, creating a \textbf{soft behavioral anchor} that naturally discourages large stylistic drift. We further strengthen this anchor by retaining only demonstration trajectories with positive total returns, ensuring that demonstration data simultaneously bias the policy toward the demonstrator's higher-value behaviors and push it toward higher proficiency. The parameter $\beta$ controls the trade-off between style fidelity and proficiency improvement.

We collect demonstration trajectories $\tau_i$ by recording the environment initialization seed and action sequences $\{a_t\}^{T}_{t=0}$. During training, complete episodes can be recovered by replaying these sequences, enabling full-episode GAE to be applied uniformly to both online and offline samples. Two types of actors are instantiated: on-policy actors using the latest policy, and LfD actors that replay recorded seeds and action sequences to generate $(s, a, r, v, adv)$ tuples. Data from both types are processed uniformly using \eqref{eq:4}. The pseudocode is shown in Algorithm~\ref{alg:algorithm}.

To quantify the similarity of the play style, we use the total variational distance, denoted $D_{policy}(\pi_1,\pi_2) = \mathbb{E}_{s\in S}\frac{1}{2}\sum_{a \in A}|\pi_1(a|s)-\pi_2(a|s)|$, which averages the total variational distance per-action over the states sampled.

\begin{algorithm}[h]
    \caption{Mixed Proximal Policy Optimization}
    \label{alg:algorithm}
    \textbf{Input}: Collections of Demonstrations: $D = \{\tau_1, \tau_2, ...\tau_N\}$,\\
    Actor policy: $\pi_\theta$, Critic policy: $V_\phi$, Demo Indicator: $d$\\
    \begin{algorithmic}[1] 
        \FOR{n=1,2,...}
        \IF {Demo Indicator $d\sim U(0,1)<\beta$}
        \STATE sample $\tau_i=\{(s_t, a_t)\}^{T}_{t=0}\sim D$ \STATE Init env with seed from $\tau_i$
        \FOR{t=0,1,...,T}
        \STATE retrieve action $a_t \in \{a_t\}^{T}_{t=0}$ from $\tau_i$
        \STATE estimate state value with $V_\phi$
        \STATE send trajectories with positive returns to learner
        \ENDFOR
        \ELSE
        \STATE Randomly initialize environment
        \FOR{t=0,1,...,T}
        \STATE sample action $a_t \sim \pi_\theta$ 
        \STATE estimate state value with $V_\phi$
        \STATE send all trajectories to learner
        \ENDFOR
        \ENDIF
        \STATE calculate advantage with GAE
        \STATE update $V_\phi$ and $\pi_\theta$ with MPPO loss \eqref{eq:4}
        \ENDFOR
    \end{algorithmic}
\end{algorithm}

\section{Experiments}
\label{others}

We aim to investigate whether MPPO can (1) meaningfully enhance agents' proficiency beyond suboptimal demonstrations and (2) retain their game styles. We adopt the IMPALA architecture~\cite{DBLP:journals/corr/abs-1802-01561} and set $\beta=0.05$ across all environments, collecting approximately 30K positive-return demonstration trajectories per demonstrator. Our seed-based storage format reduces storage by 98\%\footnote{Blackjack: 1.13MB vs 84MB, Maze: 120MB vs 18.3GB, Mahjong: 155MB vs 40GB.}. All experiments are repeated 5 times with different random seeds, as detailed in Online Resource 2.

\subsection{Blackjack}

\begin{figure}[tbhp!]
    \centering
    \includegraphics[width=\linewidth]{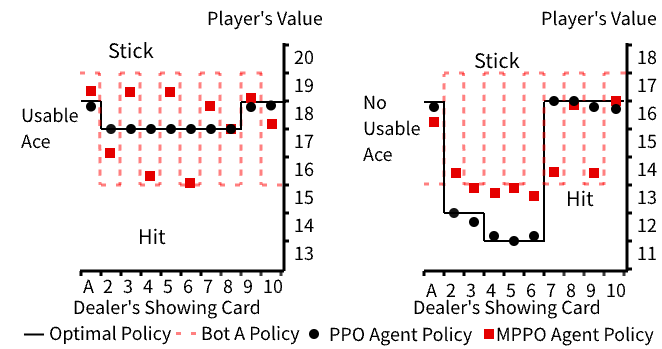}
    \caption{Decision boundaries of PPO and MPPO agents. MPPO converges between the optimal and demonstrator policies, improving proficiency while preserving style.}
    \label{fig2}
\end{figure}

Blackjack is a single-agent stochastic game where the goal is to attain a hand closer to 21 than the dealer's without exceeding this value. We provide a rule-based Bot A (red dashed lines in Figure~\ref{fig2}) and trained MPPO agents using its demonstration data.

MPPO agents achieve a win rate of 42.82\% (vs. PPO 43.40\%, optimal 43.26\%, Bot A 40.84\%), while their decision boundaries converge between the optimal and demonstrator policies (Figure~\ref{fig2}), demonstrating effective style preservation. MPPO agents remain closer to Bot A than PPO agents in policy distance ($D_{policy}=0.150$ vs.\ PPO~0.259), while PPO agents converge closer to the optimal policy ($D_{policy}=0.042$ vs.\ MPPO~0.135).

\subsection{Maze Navigation}

The maze environment is a 19$\times$19 random grid world with a guaranteed valid path (Online Resource 5). Agents observe 5 adjacent grids and move in a single direction until encountering a fork or wall. The episode ends upon reaching the exit (positive reward) or exceeding the 80-step limit. We provide two demonstrators: Bot A (right-hand wall-following) and Bot B (left-hand wall-following). Agents are evaluated on 500 unseen mazes using success rate and average steps.

\begin{table}[htbp!]
        \centering
        \caption{Maze Success Rates Results. }
        \label{tab:maze_results_wr}
        \begin{tabular}{@{}llll@{}}
        \toprule
        Agents &  Win Rate\% & \multicolumn{2}{l}{Avg Step} \\
        \midrule
        Optimal & 100.00 & \multicolumn{2}{l}{25.224}\\
        Bot A & 92.40 & \multicolumn{2}{l}{53.450}\\
        Bot B & 89.00 & \multicolumn{2}{l}{55.714}\\
        PPO  & 99.64$\pm$0.12 & \multicolumn{2}{l}{27.230$\pm$0.147}\\
        MPPO A & 99.52$\pm$0.31  &\multicolumn{2}{l}{27.649 $\pm$0.440}\\ 
        MPPO B & 99.04$\pm$0.81 & \multicolumn{2}{l}{28.104 $\pm$ 0.649}\\ 
        \bottomrule

    \end{tabular}
\end{table}

\begin{table}[htbp!]
        \centering
        \caption{Maze $D_{policy}$ Results. }
        \label{tab:maze_results_policy}
        \begin{tabular}{@{}llll@{}}
        \toprule
        Agents & Optimal &  Bot A & Bot B \\
        \midrule
        PPO  & .057$\pm$.002 &.509$\pm$.002 & .492$\pm$.001\\
        MPPO A & .084$\pm$.001 &.471$\pm$.013  &.530$\pm$.008\\ 
        MPPO B & .076$\pm$.004 &.521$\pm$.010 & .481$\pm$.009\\ 
        \bottomrule

    \end{tabular}
\end{table}

MPPO agents achieve success rates comparable to PPO while significantly outperforming their demonstrators, reducing average steps by approximately 20 (Table~\ref{tab:maze_results_wr}). Table~\ref{tab:maze_results_policy} shows MPPO maintains substantially lower $D_{policy}$ to its demonstrators than PPO. The learned strategies exhibit inherited stylistic biases: MPPO A retains right-turn navigational biases from its RHS demonstrator, while MPPO B exhibits symmetric left-turn bias. These biases produce occasional detours that account for the slightly increased path lengths in Table~\ref{tab:maze_results_wr}.

\subsection{Mahjong}
Mahjong is a multi-player imperfect-information game with approximately $10^{121}$ information sets, vastly exceeding Texas Hold'em~\cite{a16050235}. Our environment adopts the Mahjong Competition Rules (MCR) variant with 81 scoring patterns; players complete a winning hand of 14 tiles by drawing and discarding in turn (see Online Resource 1).

The MCR Mahjong bots are selected from Botzone~\cite{10.1145/3197091.3197099}. The demonstrators (Bot A, B, C) are drawn from three distinct positions on the platform's Elo ladder, and the Baseline is the top-ranked bot. 

Student agents are evaluated against the Baseline every 12 hours (512 games per evaluation). Win rates for demonstrators are from Botzone historical data. For $D_{policy}$, we collect $(s,a)$ pairs from 100 held-out trajectories and set $p(a|s)=\1_\mathrm{a=a_i}$. 
\begin{figure}[htbp]
    \centering

        \centering
        \includegraphics[width=\linewidth]{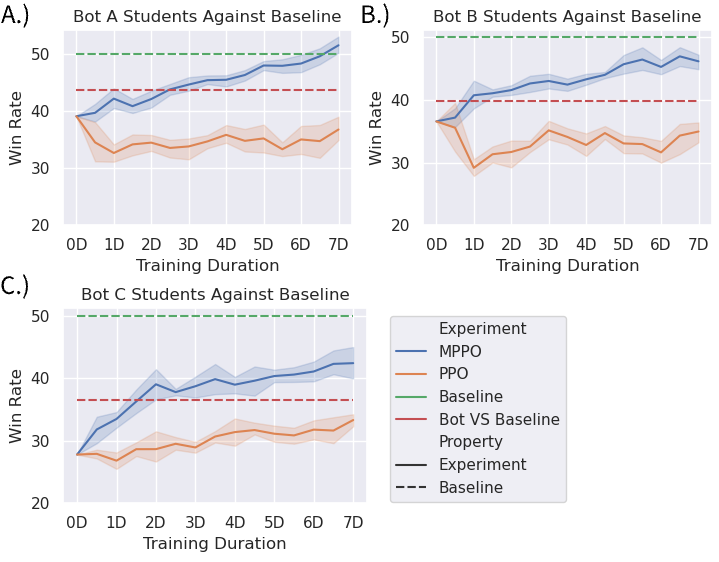} 
        \caption{Win rates against the Baseline. Red dashed lines indicate demonstrator win rates; shaded areas show 95\% confidence intervals. MPPO agents consistently surpass their demonstrators, with Bot A defeating the Baseline. To accelerate training, agents are warm-started via behavior cloning from their demonstrators (see Online Resource 3 for random initialization experiments).}
        \label{fig4}
\end{figure}

\begin{table}[htbp]
    \centering
        \centering
        \caption{Mahjong Win Rates Results.}
        \label{tab:mj_results_wr}
        \setlength{\tabcolsep}{4pt}
        \begin{tabular}{@{}lrrr@{}}
        \toprule
        Win Rate& Demonstrator& MPPO& PPO\\
        VS Base& Bot& Agents& Agents\\
        \midrule
        Bot A &43.67   & \textbf{51.05$\pm$1.43}& 36.72$\pm$3.11\\
        Bot B &39.82 & 46.17$\pm$1.72& 34.96$\pm$2.57\\
        Bot C &37.05 & 42.42$\pm$3.66& 33.32$\pm$1.41\\
        \bottomrule
    \end{tabular}
\end{table}

\begin{table}[htbp]
    \centering
        \centering
        \caption{Mahjong $D_{policy}$ Results.}
        \label{tab:mj_results_policy}
        \setlength{\tabcolsep}{4pt}
        \begin{tabular}{@{}lrr@{}}
        \toprule
        $D_{policy}$& MPPO& PPO\\
        \midrule
        Bot A & 0.297$\pm$.016& 0.678$\pm$.027\\
        Bot B & 0.318$\pm$.007& 0.691$\pm$.013\\
        Bot C & 0.279$\pm$.020& 0.772$\pm$.027\\
        \bottomrule
    \end{tabular}
\end{table}

MPPO agents quickly surpass their demonstrators; Bot A students defeat the Baseline by the end of training (Figure~\ref{fig4}, Table~\ref{tab:mj_results_wr}). Table~\ref{tab:mj_results_policy} confirms MPPO agents have significantly lower $D_{policy}$ to their demonstrators than PPO agents.

In summary, MPPO surpasses suboptimal demonstrators and, in some cases, outperforms pure PPO. $D_{policy}$ comparisons show MPPO imitates demonstrator play styles. Further ablation experiments with analysis are detailed in Online Resource 4.

\bibliographystyle{fcs}
\bibliography{ref}

\section*{MCR Mahjong Environment Description}
\label{app:env}
Mahjong is a four-player tile-based tabletop game involving imperfect information. The complexity of imperfect-information games can be quantified by information sets, which refer to game states that players are unable to differentiate based on their own observations. The average size of information sets in Mahjong is approximately $10^{48}$, rendering it a considerably more complex game to solve compared to Heads-Up Texas Hold'em, where the average size of information sets is around $10^3$. To enhance the readability of this paper, we highlight the terminologies used in Mahjong with \textbf{bold text}, and we differentiate scoring patterns with \textit{italicized text}.

In Mahjong, there are 144 tiles. Despite the existence of numerous rule variants, the general rules of Mahjong remain the same. At a broad level, Mahjong is a pattern-matching game. Each player starts with 13 tiles that are only visible to themselves, and they take turns to draw and discard one tile until a player completes a winning pattern with a 14th tile. The general winning pattern (GWP) of 14 tiles consists of four \textbf{melds} and a \textbf{pair}. A \textbf{meld} can be in the form of \textbf{Chow}, \textbf{Pung}, or \textbf{Kong}. Besides drawing all the tiles by themselves, players have the option to take the tile just discarded by another player instead of drawing one to form a \textbf{meld} or declare a win.

\paragraph{Official International Mahjong}
Official International Mahjong, also known as Mahjong Competition Rules, is a Mahjong variant aiming to enhance the game's complexity and competitiveness while weakening its gambling nature. It specifies 81 scoring patterns, which range from 1 to 88 points. In addition to forming the general winning pattern (GWP), players must accumulate at least 8 points by matching at least one scoring pattern in order to declare a win. Among the 81 patterns, 56 are highly valued and are referred to as major patterns, since most winning hands usually include at least one of them. Some special patterns do not adhere to the GWP, such as \textit{Seven Pairs}, \textit{Thirteen Orphans}, and \textit{Knitted Straight}. 

The final scores of each player depend on the winner's fan value and the provider of the 14th winning tile. Specifically, if the winner makes a winning hand of $x$ fans by drawing a tile themselves, they receive $8+x$ points from the other three players. Instead, if the 14th winning tile comes from another player, either discarded or added to an exposed pung to form a Kong, the winning player receives $8+x$ points from the provider of this tile, and only 8 points from the other two players.

\paragraph{MCR as an Environment}
As an environment, MCR exhibits several unique characteristics that pose challenges to algorithms.

First, the 8-point-to-win rule of MCR adds an additional requirement to the hand patterns. This requirement excludes many hand patterns that would otherwise be valid GWPs. Agents must be capable of distinguishing between valid and invalid hand patterns to achieve a high level of performance. In addition, the various scoring patterns of MCR render the environment multi-goaled. Although most patterns comply with the GWPs, some special patterns do not. Notably, in many situations, these special patterns can be the closest and easiest goals to pursue. These special patterns add to the diverse choices of goals other than GWPs and thus require effective exploration by agents.

Besides, the state transitions of Mahjong can be approximately represented by a directed acyclic graph. To win a game, agents are expected to make around 10 to 40 consecutive decisions. Mistakes or poorly sampled actions in Mahjong can lead to much worse game states and are hard to recover from, such as destroying some \textbf{melds}. Such a property of Mahjong conflicts with the need for exploration and poses additional challenges to learning-based agents. Furthermore, Mahjong involves high randomness and uncertainty from drawing tiles to opponent moves. During gameplay, newly encountered game states are rarely seen during training, and it is difficult and impractical to measure the similarity between states to draw on past experience. Thus, Mahjong predominantly presents out-of-distribution (OOD) states to its agents and imposes high demands on its agents' generalization capabilities. 

\paragraph{Reward Setting for MCR Mahjong Environment}
In MCR Mahjong environment, we implement dense rewards to encourage agents to approach a winning hand more quickly, by incorporating \textbf{Shanten Distance} to calculate the reward in each step. \textbf{Shanten Distance} measures the minimum distance between the agents' current hand and any valid winning pattern. Thus, agents receive a small positive reward by decreasing \textbf{Shanten Distance} and a small penalty by increasing \textbf{Shanten Distance}. 

Additionally, MCR Mahjong environment differentiates between winning by self-drawing and winning with a tile from other players. Agents will receive higher rewards if they win by self-drawing, and other players will receive the same penalty for losing. Otherwise, the winner receives a positive reward, but the player who provided the winning tile receives a larger penalty to discourage reckless play. Table~\ref{tab:mj_env} presents the reward settings for MCR Mahjong Environment.

\begin{table}[htbp]
\setlength{\tabcolsep}{2pt}

        \centering
        \caption{Reward Settings for MCR Environment}
        \label{tab:mj_env}
        
        \begin{tabular}{ll}
        \toprule
         Agent Event&  Value\\
         \midrule
         Flat Step Penalty&  -0.0006\\
         Decrease in Shanten Distance&  0.07\\
         Increase in Shanten Distance&  -0.07\\
         Win by Self-drawing&  0.8\\
         Win with Other Player's tile& 0.6\\
         Game Lost& -0.2\\
         Game Lost with playing the final tile&  -0.5\\
         Nobody Wins& 0\\
         \bottomrule
    \end{tabular}
\end{table}

\newpage

\section*{Experiment Setup and Configurations}
\label{app:config}
We conducted our experiments on Intel Xeon Gold 6348 CPU@2.6GHz platform with one Nvidia GeForce 3080 GPU and 1024GB RAM. For the software platform, we use Python 3.9.16, CUDA 12.4, Pytorch 2.5.1, and PyMahjongGB 1.2.0 on Ubuntu 20.04. 

Tables~\ref{tab:bj}, \ref{tab:maze_setting}, and \ref{tab:mcr_setting} present the experimental configurations for the Blackjack, Maze, and MCR Mahjong environments, respectively. These configurations were determined via manual parameter searching and comparative analysis of results upon agent convergence. For the MPPO students in the MCR Mahjong environment, as they continue training from behavior-cloning checkpoints, their policy networks are frozen for the first 1000 GPU iterations to fit the value networks alone without breaking the policy.

\begin{table}[htbp]
        \centering
        \caption{BlackJack Experiment Configuration}
        \label{tab:bj}
    \begin{tabular}{clcl}
    \toprule
 Entry& Setting& Entry&Setting\\
  \midrule
 Replay Buffer& & Iteration Per &\\
  -Size& 4100&  -Model Sync&1\\
         GAE Lambda&  0.98&  Entropy Coeff& 0\\
         Batch Size&  4096&  Entropy Decay& 1\\
 Policy Coeff& 1& Value Coeff&0.1\\
         Gamma&  1&  Learning Rate& 1e-2\\
         PPO Epoch&  3&  PPO Clip& 0.05\\
 Normal Actor& 76& LfD Actor&4\\
   \bottomrule
    \end{tabular}
\end{table}

\begin{table}[htbp]
\setlength{\tabcolsep}{2pt}

        \centering
        \caption{Maze Experiment Configuration}
        \label{tab:maze_setting}
        \begin{tabular}{clcl}
    \toprule
 Entry& Setting& Entry&Setting\\
  \midrule
 Replay Buffer& & Iteration Per &\\
  -Size& 8200&  -Model Sync&1\\
         GAE Lambda&  0.98&  Entropy Coeff& 0\\
         Batch Size&  8192&  Entropy Decay& 1\\
 Policy Coeff& 1& Value Coeff&0.5\\
         Gamma&  1&  Learning Rate& 5e-5\\
         PPO Epoch&  3&  PPO Clip& 0.05\\
 Normal Actor& 75& LfD Actor&5\\
   \bottomrule
    \end{tabular}
    \end{table}
\begin{table}[htbp]
        \centering
        \caption{MCR Mahjong Experiment Configuration}
        \label{tab:mcr_setting}
    \begin{tabular}{clcl}
    \toprule
 Entry& Setting& Entry&Setting\\
  \midrule
 Replay Buffer& & Iteration Per &\\
 -Size& 4100&  -Model Sync&1\\
         GAE Lambda&  0.98&  Entropy Coeff& 1.5e-1\\
         Batch Size&  4096&  Entropy Decay& 0.99998\\
 Policy Coeff& 1& Value Coeff&1\\
         Gamma&  1&  Learning Rate& 1e-5\\
         PPO Epoch&  5&  PPO Clip& 0.05\\
 Normal Actor& 70& LfD Actor&10\\
   \bottomrule
    \end{tabular}
\end{table}

\newpage
\begin{table}[htbp]
        \centering

    \begin{tabular}{clcl}
 \\
 \\
    \end{tabular}
\end{table}

\newpage

\section*{Mahjong MPPO Agent with Random Initialization}
\label{app:rand_init}
In the main experiments, MPPO and PPO agents in the Mahjong environment were warm-started from behavior cloning (BC) checkpoints derived from their respective demonstrator bots. We clarify that this warm-start is employed solely to reduce wall-clock time and resource consumption and is not a methodological prerequisite for MPPO.

To validate this claim, we conducted an ablation study on the Mahjong environment using Bot A's demonstration data. We compared an MPPO agent initialized from a BC checkpoint against an MPPO agent initialized from random weights. All other experimental settings, including network architecture, hyperparameters, and the mixture ratio of demonstration data ($\beta=0.05$), remained identical.

\begin{figure}[tbhp!]
    \centering
        \centering
        \includegraphics[width=\linewidth]{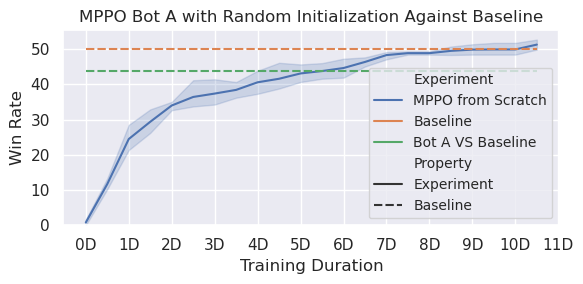} 
        \caption{Win Rate Against the Baseline for MPPO Agent A When Trained from Random Initialization.}
        \label{fig:mj_rand_init_wr}
\end{figure}

\begin{figure}[tbhp!]
        \centering
        \includegraphics[width=\linewidth]{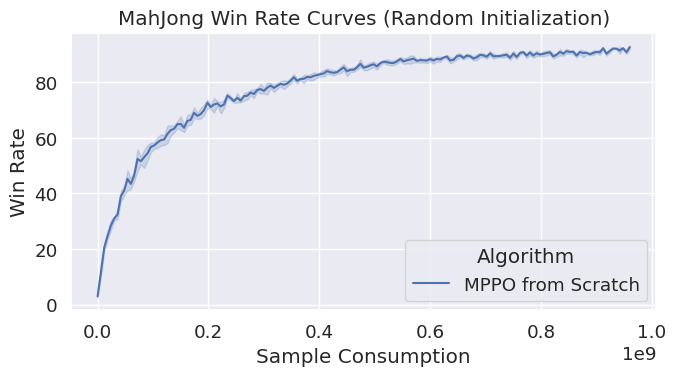}
        \caption{Win Rate Curves During Training for MPPO Agent A (Random Initialization).}
        \label{fig:mj_rand_init_curves}

\end{figure}

The final win rates were nearly identical, and the policy distance to the demonstrator remained low in both cases, confirming that the demonstration data integrated via the MPPO objective is sufficient to effectively guide policy improvement and style preservation, even without a pre-trained policy. The complete win rate curve against the baseline bot and the corresponding training curve for this experiment are presented in Figure~\ref{fig:mj_rand_init_wr} and Figure~\ref{fig:mj_rand_init_curves}, respectively.

\newpage
\begin{table}[htbp]
        \centering

    \begin{tabular}{clcl}
 \\
 \\
    \end{tabular}
\end{table}

\newpage

\section*{Ablation and Comparative Study}
\label{app:ablation}

\begin{table}[ht!]
\caption{Summarized ablation and comparison study results for A) Win rates and B) $D_{policy}$  with each environment's Bot A. Ablation and comparative study results are separated by a horizontal line. Highest Win rates and lowest $D_{policy}$ values in ablation study results are highlighted.}
    \setlength{\tabcolsep}{4pt}
    \centering
    \begin{tabular}{@{}l@{}lrrr@{}}
        \toprule
        A)&Win Rate $\%$& Blackjack& Maze& Mahjong\\
        \midrule
        &PPO &43.40$\pm$0.17& \textbf{99.64$\pm$0.12}& 36.72$\pm$3.11\\
        &MPPO Ref &42.82$\pm$0.08& 99.52$\pm$0.31& \textbf{51.05$\pm$1.43}\\
        &2x Demo&42.08$\pm$0.06& 99.60$\pm$0.20& 48.09$\pm$1.86\\
        &0.5x Demo&42.93$\pm$0.12& 99.12$\pm$0.99& 46.84$\pm$1.97\\
        &All Data&\textbf{43.62$\pm$0.04}& 94.24$\pm$0.81& 42.30$\pm$3.65\\
        &TD(0) Adv&43.31$\pm$0.28& 99.40$\pm$0.69& 17.34$\pm$0.74\\
        \midrule
        &GAIL&25.63$\pm$1.21& 92.40$\pm$0.00& 3.32$\pm$0.58\\
        &SAIL&38.50$\pm$0.01& 94.52$\pm$0.70& 19.73$\pm$3.30\\
        &DQfD&42.26$\pm$0.29& 87.20$\pm$2.62& 15.43$\pm$1.38\\
 & PPOfD& 42.89$\pm$0.15& 99.40$\pm$0.20&41.52$\pm$2.59\\
        \midrule
        B)&$D_{policy}$& Blackjack& Maze& Mahjong\\
        \midrule
        &PPO &0.259$\pm$.006& 0.509$\pm$.002& 0.678$\pm$.027\\
        &MPPO Ref &0.150$\pm$.011& 0.471$\pm$.013& 0.297$\pm$.016\\
        &2x Demo&\textbf{0.093$\pm$.007}& \textbf{0.457$\pm$.018}& \textbf{0.287$\pm$.006}\\
        &0.5x Demo&0.186$\pm$.003& 0.492$\pm$.012& 0.324$\pm$.007\\
        &All Data&0.294$\pm$.003& 0.572$\pm$.016& 0.775$\pm$.057\\
        &TD(0) Adv&0.218$\pm$.004& 0.523$\pm$.003& 0.727$\pm$.042\\
        \midrule
        &GAIL&0.501$\pm$.010& $2.1\mathrm{e}{-6}\pm0$& 0.784$\pm$.044\\
        &SAIL&0.495$\pm$.006& 0.540$\pm$.017& 0.690$\pm$.025\\
        &DQfD&0.411$\pm$.003& 0.659$\pm$.001& 0.793$\pm$.004\\
 & PPOfD& 0.231$\pm$.011& 0.347$\pm$.009&0.335$\pm$.019\\
 \bottomrule
    \end{tabular}
    \label{tab:ablation_study}
\end{table}

To analyze the impact of different components of the MPPO algorithm, we conducted ablation studies using Bot A's trajectories across each environment. Table~\ref{tab:ablation_study} summarizes the win rates and $D_{policy}$ values between the agent groups and their corresponding demonstrators.

For the \textbf{2x Demo} and \textbf{0.5x Demo} experiments, we doubled and halved the value of $\beta$, respectively, to analyze the impact of the demonstration data ratio. As expected, a higher ratio of demonstration data leads to a lower $D_{policy}$. The proportion of demonstration data also affects the final proficiency of the agents. In each environment, the proficiency metrics of MPPO agents peak at different ratios, indicating that different environments correspond to unique optimal ratios of demonstration data.

For the \textbf{All Data} experiments, we regenerate all datasets to include all trajectories without filtering by positive returns. Without this filtering, the demonstration data includes low-quality or negative-return trajectories that provide no meaningful behavioral guidance, reducing the entire algorithm to online PPO where a fraction of the actors sample from fixed seed environments with a fixed policy. In this setting, $D_{policy}$ values are high in all environments, and the win rates vary by environment.

For the \textbf{TD(0) Adv} experiments, we replace GAE with 1-step TD advantage, an approach commonly adopted in existing offline RL and LfD methods. Because 1-step TD responds slowly to the final reward, the advantage estimates for individual actions become less reliable, weakening the implicit behavioral anchoring effect of the filtered demonstration data. Consequently, we observe higher $D_{policy}$ values in all settings. While TD(0) performs well in Blackjack and Maze, it struggles in Mahjong, a more complex environment with long-horizon decision sequences, as it fails to leverage all future information.

For the comparative study, we compare MPPO with other LfD and IL methods: GAIL, SAIL, and DQfD. As shown in Table~\ref{tab:ablation_study}A, GAIL and SAIL perform well in Maze, a 2D state-space environment, yet struggle in Mahjong, where defining similarity between state-action pairs is challenging — a known limitation of adversarial IL methods in high-dimensional settings.

DQfD also performs poorly in Mahjong: its training trajectories for the game exhibit the same low-entropy characteristics as those of the MPPO algorithm. To eliminate the influence of backbone algorithms and differences in action-sampling strategies, we ported DQfD's explicit supervised loss to MPPO, creating \textit{PPOfD}. In essence, PPOfD differs from MPPO solely in the mechanism by which it encourages students to imitate demonstrators. PPOfD outperforms DQfD across all environments; it is comparable to MPPO in Blackjack and Maze but lags significantly behind MPPO in Mahjong. This indicates that our implicit behavior cloning constraint is more adaptable to diverse environments than explicit loss functions.

Regarding play styles, as shown in Table~\ref{tab:ablation_study}B, MPPO is the only method that meaningfully maintains low $D_{policy}$ values while improving agent proficiency across all environments.

\newpage

\section*{Maze Environment Illustration}
\label{app:maze_fig}

\begin{figure}[htbp!]
    \centering
    \includegraphics[width=0.8\linewidth]{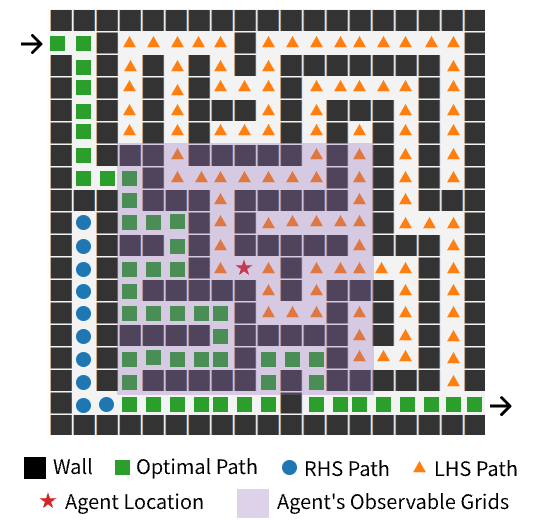}
    \caption{An example of a $19\times19$ random maze used in the navigation task. The agent observes the 5 adjacent grids around its current position (highlighted). The environment guarantees a valid path from the entrance to the exit.}
    \label{fig3}
\end{figure}

\end{document}